\PassOptionsToPackage{table}{xcolor}
\documentclass{article}
\usepackage{arxiv}

\usepackage{amsmath,amssymb,mathtools}
\usepackage{booktabs,multirow,array,tabularx,colortbl}
\usepackage{graphicx,subcaption}
\usepackage{xcolor}
\definecolor{fwd}{HTML}{3E6F9E}
\definecolor{bwd}{HTML}{6E5F8C}
\definecolor{inv}{HTML}{B4585A}
\definecolor{act}{HTML}{B07A2E}
\definecolor{ours}{HTML}{598BE7}
\definecolor{oursbg}{HTML}{E7EFFC}

\makeatletter
\let\NAT@parse\undefined
\makeatother
\makeatletter
\renewcommand\figure{\@ifnextchar[\@figfps{\@float{figure}[tbp]}}
\def\@figfps[#1]{\@float{figure}[tbp]}
\def\@tabskips{%
  \setlength{\abovecaptionskip}{\@belowcaptionskip}%
  \setlength{\belowcaptionskip}{\@abovecaptionskip}}
\renewcommand\table{\@ifnextchar[\@tabfps{\@tabskips\@float{table}[tbp]}}
\def\@tabfps[#1]{\@tabskips\@float{table}[tbp]}
\makeatother

\usepackage[numbers,sort&compress]{natbib}
\usepackage{hyperref,url}
\hypersetup{colorlinks=true, linkcolor=ours, citecolor=ours, urlcolor=ours, filecolor=ours}
\hypersetup{pdftitle={Sigma-LeWM: Learning World Models via S-Information for Robot Control},
            pdfauthor={Xinrui Zu and Michael Przystupa and Zhao Yang and Shujian Yu and Kevin Sebastian Luck}}

\usepackage{enumitem}
\usepackage{algorithm,algpseudocode}
\usepackage{microtype}
\graphicspath{{./}{figures/}}

\newcommand{\std}[1]{{\scriptsize\color{gray}$\pm$#1}}

\newif\ifreviewmark
\reviewmarkfalse
\newcommand{\revmark}{\ifreviewmark\color{blue}\fi}
\newcommand{\revcap}{\ifreviewmark\captionsetup{font+=revblue}\fi}%
\newcommand{\revframe}[1]{\ifreviewmark{\setlength{\fboxsep}{1.5pt}\setlength{\fboxrule}{0.6pt}\fcolorbox{blue}{white}{#1}}\else#1\fi}
\newcommand{\ours}{$\Sigma$-LeWM}

\newcommand{\LB}{\mathcal{L}_{\mathrm{back.}}}
\newcommand{\LA}{\mathcal{L}_{\mathrm{inv.}}}

\newcommand{\enc}{E}

\usepackage[most]{tcolorbox}
\tcbset{
  thmcard/.style={
    enhanced, breakable, rounded corners, arc=3.5pt, boxrule=0.5pt,
    colback=white, colframe=ours!42!white,
    left=5pt, right=5pt, top=4pt, bottom=4pt,
    before skip=6pt, after skip=6pt,
    borderline west={1.4pt}{0pt}{ours!70!white},
  }
}

\newcommand{\figorph}[3]{\IfFileExists{#1}{\includegraphics[width=#2]{#1}}{\fbox{\parbox{\dimexpr#2-2\fboxsep-2\fboxrule}{\centering\vspace{#3}\small\textsc{pending figure}\\ \texttt{\detokenize{#1}}\vspace{#3}}}}}

\usepackage{caption}
\DeclareCaptionFont{revblue}{\color{blue}}%
\usepackage{etoolbox}

\newlength{\floatcolw}
\AtBeginDocument{%
  \setlength{\floatcolw}{0.78\textwidth}%
  \captionsetup{width=\floatcolw}%
}
\makeatletter
\g@addto@macro\@floatboxreset{%
  \setlength{\columnwidth}{\floatcolw}%
  \setlength{\linewidth}{\floatcolw}%
  \centering
}
\makeatother

\renewcommand{\headeright}{Preprint}
\renewcommand{\undertitle}{Preprint}
\renewcommand{\shorttitle}{$\Sigma$-LeWM}

\usetikzlibrary{svg.path}
\definecolor{oursink}{HTML}{2F5FB8}%
\newcommand{\icoglobe}{\tikz[baseline=-0.62ex,line width=0.55pt,line cap=round,oursink]{%
  \draw (0,0) circle (0.78ex);
  \draw (0,0) ellipse (0.34ex and 0.78ex);
  \draw (-0.78ex,0)--(0.78ex,0);
  \draw (-0.67ex,0.40ex)--(0.67ex,0.40ex);
  \draw (-0.67ex,-0.40ex)--(0.67ex,-0.40ex);}}
\newcommand{\icohf}{\tikz[baseline=-0.62ex, x=0.0867ex, y=-0.0867ex,
  line width=0.55pt, line cap=round, line join=round, oursink]{%
  \draw (0,0) circle (9);
  \draw (-3.5,-1.5)--(-3.5,-2);
  \draw (3.5,-1.5)--(3.5,-2);
  \draw (-4,2) .. controls (-3,3.8) and (-1.6,4.6) .. (0,4.6)
               .. controls (1.6,4.6) and (3,3.8) .. (4,2);}}
\newcommand{\icogithub}{\raisebox{\dimexpr\depth-0.3ex\relax}{\scalebox{0.118}[-0.118]{\resizebox{!}{16ex}{%
  \tikz[oursink]{\fill svg{M 8 0 C 3.58 0 0 3.58 0 8 c 0 3.54 2.29 6.53 5.47 7.59 c 0.4 0.07 0.55 -0.17 0.55 -0.38 c 0 -0.19 -0.01 -0.82 -0.01 -1.49 c -2.01 0.37 -2.53 -0.49 -2.69 -0.94 c -0.09 -0.23 -0.48 -0.94 -0.82 -1.13 c -0.28 -0.15 -0.68 -0.52 -0.01 -0.53 c 0.63 -0.01 1.08 0.58 1.23 0.82 c 0.72 1.21 1.87 0.87 2.33 0.66 c 0.07 -0.52 0.28 -0.87 0.51 -1.07 c -1.78 -0.2 -3.64 -0.89 -3.64 -3.95 c 0 -0.87 0.31 -1.59 0.82 -2.15 c -0.08 -0.2 -0.36 -1.02 0.08 -2.12 c 0 0 0.67 -0.21 2.2 0.82 c 0.64 -0.18 1.32 -0.27 2 -0.27 c 0.68 0 1.36 0.09 2 0.27 c 1.53 -1.04 2.2 -0.82 2.2 -0.82 c 0.44 1.1 0.16 1.92 0.08 2.12 c 0.51 0.56 0.82 1.27 0.82 2.15 c 0 3.07 -1.87 3.75 -3.65 3.95 c 0.29 0.25 0.54 0.73 0.54 1.48 c 0 1.07 -0.01 1.93 -0.01 2.2 c 0 0.21 0.15 0.46 0.55 0.38 A 8.013 8.013 0 0 0 16 8 c 0 -4.42 -3.58 -8 -8 -8 z};}}}}}
\newcommand{\btnfont}{\fontfamily{DejaVuSansMono-TLF}\selectfont}
\newtcbox{\linkpill}{on line, arc=3.2pt, boxrule=0.4pt, colframe=ours!32, colback=oursbg,
  left=6.5pt, right=7.5pt, top=2.6pt, bottom=2.6pt, boxsep=0pt}
\newcommand{\linkbutton}[3]{\href{#1}{\linkpill{#2\hspace{0.45em}%
  \textcolor{oursink}{\btnfont\small #3}}}}
\newcommand{\websiteurl}{https://zuxinrui.github.io/sigma-lewm/}
\newcommand{\codeurl}{https://github.com/zuxinrui/sigma-lewm}
\newcommand{\hfurl}{https://huggingface.co/zuxinrui/sigma-lewm}

\makeatletter
\renewcommand{\@maketitle}{%
  \vbox{%
    \hsize\textwidth \linewidth\hsize
    \vskip 0.1in
    \hrule height 4\p@ \vskip 0.25in \vskip -\parskip
    \centering
    {\LARGE\bfseries \@title\par}
    \vskip 0.29in \vskip -\parskip \hrule height 1\p@ \vskip 0.24in
    {\bfseries \authornames\par}
    \vskip 5\p@
    {\authoraffil\par}
    \vskip 0.17in
    {\linkbutton{\websiteurl}{\icoglobe}{Website}\hspace{5pt}%
     \linkbutton{\codeurl}{\icogithub}{Code}\hspace{5pt}%
     \linkbutton{\hfurl}{\icohf}{Hugging Face}\par}
    \vskip 0.16in
  }%
}
\makeatother

\title{$\Sigma$-LeWM: Learning World Models via S-Information\\ for Robot Control}
\newcommand{\authornames}{Xinrui Zu \quad Michael Przystupa \quad Zhao Yang \quad Shujian Yu \quad Kevin Sebastian Luck}
\newcommand{\authoraffil}{Vrije Universiteit Amsterdam, The Netherlands}
\author{\authornames}
\date{}

\begin{document}

\maketitle
\begin{abstract}
World models have increasingly been applied as a model-based control method for robotics tasks such as manipulation and locomotion. While early world model architectures relied on decoders to prevent mode collapse and learn a predictable latent space, recent works on world models, such as Joint Embedding Predictive Architecture or LeWorldModel, increasingly aim to learn embeddings without a grounding decoder, utilising only a latent forward dynamics predictor and a stabilising entropy term. In this paper, we set out to demonstrate that the further introduction of two additional latent losses to LeWorldModel, a backward dynamics predictor and an inverse dynamics action predictor, enforces further structure on the latent embedding, leading to further gains in both performance and planning efficiency. These are critical properties for improving the applicability of such methods to real-time model-based control of real-world robots. We demonstrate these gains on standard world model benchmarks and two simulated robot manipulation tasks. 

\end{abstract}

\vfill
\begin{center}
    \begin{minipage}{\linewidth}\centering
    \includegraphics[width=0.65\linewidth]{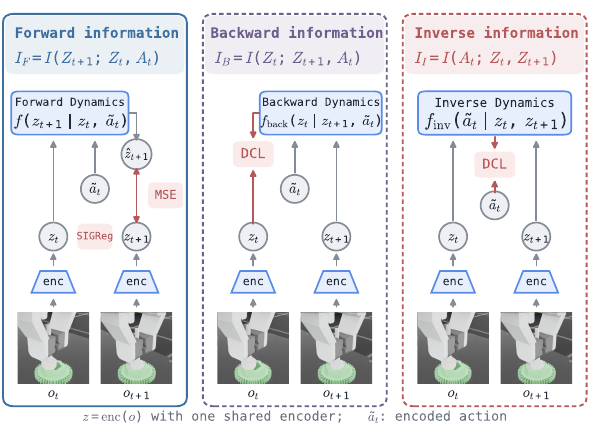}
    \captionof{figure}{$\Sigma$-LeWM introduces two auxiliary tasks for the latent dynamics learning in LeWorldModel (LeWM): Next to learning the standard forward dynamics function in the latent space (left), $\Sigma$-LeWM learns also implicitly a backward dynamics (middle) and inverse dynamics function (right). Learning these functions alongside standard LeWM imposes further restrictions and biases on the latent embedding space. At test time actions are found by standard forward planning with the forward dynamics (left) and model-predictive control (MPC) algorithms.}
    \label{fig:method-overview}
    \end{minipage}
\end{center}
\vfill
\clearpage
\section{Introduction}

World models have become increasingly useful for model-based control of robots operating in complex observation spaces such as images or high-dimensional proprioceptive data~\citep{Finn2017DVF,lancaster2024modemv2,ye2026world,zhu2025unified,huang2025point2plans}. Instead of planning directly in complex high-dimensional spaces, the core idea is to learn an embedding space of observations and perform planning in a dimensionally reduced space efficiently and quickly~\citep{Hafner2019}. This requires an embedding function that projects high-dimensional observations into latent features where a transition function can be learned to perform model-predictive control. In this work, we investigate ways to improve world model learning by incorporating multiple learning signals during optimisation.

Early world models have relied heavily on encoder-decoder architectures, such as variational autoencoders~\citep{kingma2013auto}, to prevent mode collapse and ensure reliable learning for dynamics learning and planning. The decoder's prediction of the original observation typically imposes artificial restrictions on the learned latent space: each observation feature must be fully predictable to satisfy and minimise the reconstruction loss. Instead, in the context of planning and model-based control, our main objective is a structured latent space that places similar observations close together, while maximising the ability of a learned transition function to predict the next latent state given an action.

As prior work has shown, removing the decoder, while challenging, yields better-structured latent spaces and improved planability, leading to performance gains in robot control tasks~\citep{pmlr-v162-hansen22a}. Notably, recent work such as LeWorldModel (LeWM)~\citep{Maes2026} shows that using only a latent transition loss that predicts the next latent features, along with a Sketched-Isotropic-Gaussian Regularizer to prevent mode collapse in latent representations, enables training world models applicable to robotic tasks such as object manipulation.
In this work, we go one step further: we show a natural and intuitive extension to LeWorldModel, motivated by insights from information theory, in which an additional backward transition-dynamics loss and an inverse dynamics loss induce further structure on the world model's latent embedding space. We show that this leads not only to better performance, but also to computational efficiency by allowing reductions in both horizon length and inference steps, which lowers the test-time planning compute of the model-predictive control process and better real-world applicability in robotics.

In summary, the contributions of our paper are:
\begin{itemize}
    \item[C1] We propose a new extension of LeWM, namely $\Sigma$-LeWM, which introduces backward and inverse dynamics as auxiliary losses/tasks to learn a more structured \& predictable latent embedding.
    \item[C2] We show that this approach has an intuitive grounding in information theory, where this learning approach is motivated by maximising the S-Information.
    \item[C3] Our experiments demonstrate that $\Sigma$-LeWM leads to better performance than standard LeWM, and that it also does so on more realistic robotics tasks such as gear assembly.
\end{itemize}

\section{Related Works}
\label{sec:related}

\textbf{Latent World Models and JEPA.}
Generative world models like the Dreamer family~\citep{Hafner2019}, PlaNet~\citep{pmlr-v97-hafner19a}, and DIAMOND~\citep{Alonso2024} learn latent dynamics through reconstruction with decoders and reward signals.
TD-MPC~\citep{pmlr-v162-hansen22a}, instead, drops the decoder and learns the latent dynamics via temporal-difference objectives.
Recently, JEPA~\citep{Assran_2023_CVPR} has gained increasing attention with no need for reconstruction or reward, and learns to predict latent representations using an encoder and a predictor.
Methods like I-JEPA~\citep{Assran_2023_CVPR} and V-JEPA~\citep{assran2025v} use JEPA for self-supervised representation learning by predicting the latent representation of masked inputs.
However, latent prediction does not guarantee informative representations and can suffer from representational collapse.

\textbf{Anti-collapse in Latent World Models.}
Recent JEPA-like methods use various approaches to alleviate representation collapse and stabilise training: DINO-WM~\citep{pmlr-v267-zhou25t} freezes a pretrained encoder to prevent the prediction from collapsing; while PLDM~\citep{Sobal2025}, Sub-JEPA~\citep{Zhao2026}, and RC-aux~\citep{Li2026} use auxiliary objectives to regularise the latent space.
However, preventing collapse alone does not guarantee that the latent space captures control-relevant structure, which is useful in robot learning and planning.

\textbf{Inverse Dynamics in Latent World Models.}
Recent works increasingly use inverse dynamics to make latent representations more control-relevant.
SMWM~\citep{Ivashkov2026} and AC-MTM~\citep{Boylan2026} use inverse action prediction from the two latent representations as an auxiliary task when learning representations.
On the other hand, GC-IDM~\citep{Nguyen2026} and ACID~\citep{Seo2026} use inverse dynamics only for downstream planning, without changing the latent representation learning.

\section{Preliminaries}
\subsection{LeWorldModel}
We assume a decoder-less latent world model consisting of 
\begin{align}
    \text{an Encoder } & \enc:o_t\mapsto z_t, \\
    \text{latent transition function } & f:z_t, a_t \mapsto z_{t+1},
\end{align}
where the variables $z_t$ and $z_{t+1}$ represent latent encodings of observations $o_t$ and $o_{t+1}$ respectively. 
Observations $o_t$ are often assumed to be high-dimensional, such as images.
In LeWM, the functions given above are trained with the two objectives combining the latent next-state prediction term $\mathcal L_{\mathrm{pred}}$ and a latent state regularisation term $\mathcal L_{\mathrm{reg}}$, i.e.,
\begin{equation}
    \mathcal L_{\mathrm{LeWM}}
    =
    \underbrace{
    \|f(z_t,a_t)-z_{t+1}\|_2^2
    }_{\mathcal L_{\mathrm{pred}}}
    +
    \lambda_{\mathrm{sig}}
    \underbrace{
    \mathcal L_{\mathrm{SIGReg}}(Z)
    }_{\mathcal L_{\mathrm{reg}}},
    \label{eq:lewm-objective}
\end{equation}
where SIGReg is a regularising term introduced by LeWM to prevent mode collapse of the latent embedding in the absence of a decoder. 

At test time, a planning algorithm in the latent space of the world model computes the actions. 
Given the representation of the current state $\enc(o_0)=z_0$ and an encoded goal state $\enc(o_\text{goal}) = z_\text{goal}$, an MPC-like planning method samples candidate action sequences and rolls them forward in the latent embedding space. 
Thus, the planner optimises the objective 
\begin{equation}
    a_{0:H-1}^{\star}
    =
    \arg\min_{a_{0:H-1}}
    \left\|
        h(z_0,a_{0:H-1},f)-z_{\mathrm{goal}}
    \right\|_2^2,
    \label{eq:latent-planning}
\end{equation}
where $h(z_0,a_{0:H-1},f) = \Tilde{z}_{H}$ is the predicted final latent state of a latent trajectory with start state $z_0$ and rolled forward using the learned transition function $z_{t+1} = f(z_t, a_t)$. 
Following LeWM, we also use the cross-entropy method (CEM), a sampling-based optimiser that iteratively selects candidate action sequences according to their latent-space goal cost, to find the optimal trajectory. 
Once an optimal latent trajectory is found, the first action(s) is executed on the system before the next planning cycle starts.

\subsection{Decoupled Contrastive Information Maximisation}

To optimise the backward and inverse predictive objectives, we adopt the decoupled contrastive loss (DCL)~\citep{Yeh2022}. DCL can be viewed as a contrastive surrogate for mutual information maximisation, closely related to the InfoNCE lower bound~\citep{Oord2018,pmlr-v97-poole19a}.

For a batch of paired representations
$\{(q_i,k_i)\}_{i=1}^{N}$, we define the normalised similarity $s_{ij}
= \frac{\langle \hat q_i,\hat k_j\rangle}{\tau}$,
where $\hat q_i = q_i/\|q_i\|_2$ and $\hat k_j = k_j/\|k_j\|_2$ are the normalised critic outputs, and $\tau$ is a temperature parameter.

The directional DCL objective is defined as:
\begin{equation}
\mathcal{L}_{\mathrm{DCL}}^{q\rightarrow k}
=
\frac{1}{N}\sum_i
\left[
-s_{ii}
+
\log\sum_{j\neq i}\exp(s_{ij})
\right].
\end{equation}
We use its symmetric form
\begin{equation}
\mathcal{D}(Q,K)
=
\frac{1}{2}
\left(
\mathcal{L}_{\mathrm{DCL}}^{q\rightarrow k}
+
\mathcal{L}_{\mathrm{DCL}}^{k\rightarrow q}
\right).
\label{eq:dcl}
\end{equation}

Unlike InfoNCE, DCL removes the positive pair from the
normalisation term, thereby decoupling positive attraction from negative-sample repulsion. For each sample, the corresponding InfoNCE loss satisfies:
\begin{equation}
\ell_{\mathrm{InfoNCE}}
=
\log\left(1+\exp(\ell_{\mathrm{DCL}})\right),
\end{equation}
which is a strictly increasing function of
$\ell_{\mathrm{DCL}}$. Therefore, minimising DCL is consistent with maximising the corresponding InfoNCE mutual-information lower bound.

\section{Methodology}
In this section, we first introduce our proposed $\Sigma$-LeWM, an extension of the standard LeWM, and then provide formal intuition for the effects of the proposed architecture.

\subsection{Overview of \ours}
We extend the standard latent world model, specifically LeWM, with two additional predictive functions: the backward and inverse transition functions. \\
\textbf{Components:} Therefore, our proposed extension to LeWM contains the following main components:
\begin{subequations}
\label{eq:world_model_components}
\begin{align}
&\text{Observation encoder: } 
& \mathbf{z_t} &= \enc^O(o_t) \label{eq:context_encoder} \\
&\text{Action encoder:}
& \Tilde{\mathbf{a}}_t &= \enc^A(a_t) \\
&\text{Forward Dynamics:}
& \mathbf{z}_{t+1} &= f(\mathbf{z}_t, \Tilde{\mathbf{a}_t}) \\
&\text{Backward Dynamics (aux.):}
& \mathbf{z}_t &= f_\text{back}(\mathbf{z}_{t+1}, \Tilde{\mathbf{a}_t}) \\
&\text{Inverse Dynamics (aux.):}
& \Tilde{\mathbf{a}}_t &=  f_\text{inv}(\mathbf{z}_{t}, \mathbf{z}_{t+1}) 
\end{align}
\end{subequations}

Building on LeWM, $\Sigma$-LeWM retains the standard observation and action encoders $\enc(\cdot)$ and $\enc^A(\cdot)$, together with the forward dynamics model $f(\mathbf{z}_t,\tilde{\mathbf a}_t)$.
These auxiliary objectives ($f_\text{back.}(\cdot)$ and $f_\text{inv.}(\cdot)$) are used only during training and do not introduce an additional dynamic model at inference time. 
They can be trained implicitly by embedding them directly in the DCL loss (see Sec.~\ref{sec:method_training}). 
Consequently, test-time planning remains identical to LeWM. Nevertheless, by maximising the backward and inverse information objectives during training, $\Sigma$-LeWM imposes additional structural biases on the learned latent space.

\subsection{Training}\label{sec:method_training}
To train the individual functions of our world model in Eq~(\ref{eq:world_model_components}), we use an objective composed of three individual loss terms, namely
\begin{equation}
\label{eq:objective}
\mathcal L_{\Sigma}
= \mathcal{L}_{\mathrm{forward}}
+
\lambda_{\mathrm{back.}}\LB
+
\lambda_{\mathrm{inv.}}\LA ,
\end{equation}
where $\mathcal L_{\mathrm{forward}}$ is the forward world-model objective and $\LB,\LA$ optimise the backward and inverse views, respectively. 
$\lambda_{\mathrm{back}}$ and $\lambda_{\mathrm{inv.}}$ are regularisation weights. 
\newline 
\textbf{Forward Loss ($\mathcal L_{\mathrm{forward}})$:} The forward network is trained with the same loss functions as LeWM plus a multi-step prediction term with weight $\lambda_{\mathrm{multi}}=1$, i.e., $\mathcal L_{\mathrm{forward}} = \mathcal L_{\mathrm{LeWM}} + \sum_{k=2}^{m} \| \hat z_{t+k} - z_{t+k} \|_2^2,$ together with LeWM's latent space regularisation term.
\newline 
\textbf{Backward Loss ($\mathcal L_{\mathrm{back}})$:} To train the backward dynamics function $\mathbf{z}_t = f_\text{back}(\mathbf{z}_{t+1}, \Tilde{\mathbf{a}_t})$ we utilise the DCL as introduced in Eq.~(\ref{eq:dcl}). 
Given a transition $(o_t, a_t, o_{t+1})$ we define the backward loss as:
\begin{equation}
    \mathcal{L}_{\text{back.}} = \mathcal{D}(f_{\text{back.}}(z_{t+1}, \Tilde{a}_t), z_{t})
\end{equation}
where $z_{t+1} = \enc(o_{t+1})$, $\Tilde{a}_t = \enc^A(a_t)$ and $z_t = \enc({o_t})$ with the goal of aligning the representation of the future state-action pair $(z_{t+1},\tilde a_t)$ with the latent representation $z_t$ of the current observation $o_t$. 
In our experiments, we found that training an additional encoder in the representation $z_t$ improved performance; therefore, in our implementations, we optimise $\mathcal{L}_{\text{back.}} = D(f_{\text{back.}}(z_{t+1}, \Tilde{a}_t), E^z(z_{t}))$. 
\newline 
\textbf{Inverse Loss ($\LA$):} In a similar fashion to the backward loss, we utilise the DCL loss in Eq.~(\ref{eq:dcl}) again for training the inverse dynamics function with:
\begin{equation}
    \mathcal{L}_{\text{inv.}} = \mathcal{D}(f_{\text{inv.}}(z_t,z_{t+1}), \tilde{a}_{t}).
\end{equation}
This objective maximises the inverse information dependence between the action representation $\tilde a_t$ and the pair of consecutive latent states $(z_t, z_{t+1})$.

\subsection{Theoretical Analysis: An Information-Theoretic Intuition}
An information-theoretic view of the above learning objectives provides a unified interpretation of the forward, backward, and inverse objectives through multivariate information maximisation.

For the following discussion, we will use $Z_t, Z_{t+1}$ and $\Tilde{A}_t$ for readability and conciseness, but note that these are direct encodings of the original observations $O_t, O_{t+1}$ and actions $A_t$. 
{
The objective of LeWM in Eq.~(\ref{eq:lewm-objective}) contains two terms. Regarding the prediction term, let us assume a fixed-variance Gaussian predictive model:
\begin{equation}
    p_{\phi}(Z_{t+1}\mid Z_t,A_t)
    =
    \mathcal{N}
    \left(
    g_{\phi}(Z_t,A_t),\sigma^2 I
    \right).
\end{equation}
We have:
\begin{align}
-\log p_{\phi}(Z_{t+1}\mid Z_t,A_t)
& =
\frac{1}{2\sigma^2}
\left\|Z_{t+1}-g_{\phi}(Z_t,A_t)\right\|_2^2 \nonumber \\
&\quad +\frac{d}{2}\log(2\pi\sigma^2),
\end{align}
hence minimising the prediction MSE is equivalent, up to an additive constant, to minimising the conditional negative log-likelihood, which can be further decomposed as:
\begin{align}
&\mathbb{E}_{p(Z_t,A_t,Z_{t+1})}
\!\left[
-\log p_{\phi}(Z_{t+1}\mid Z_t,A_t)
\right]
\nonumber\\
&=
\mathbb{E}_{p(Z_t,A_t,Z_{t+1})}
\!\left[
-\log p(Z_{t+1}\mid Z_t,A_t)
\right]
\nonumber\\
&\quad+
\mathbb{E}_{p(Z_t,A_t,Z_{t+1})}
\!\left[
\log
\frac{
p(Z_{t+1}\mid Z_t,A_t)
}{
p_{\phi}(Z_{t+1}\mid Z_t,A_t)
}
\right]
\nonumber\\
&=
H(Z_{t+1}\mid Z_t,A_t)
\nonumber\\
&\quad+
\mathbb{E}_{p(Z_t,A_t)}
D_{\mathrm{KL}}\!\left(
p(Z_{t+1}\mid Z_t,A_t)
\|
p_{\phi}(Z_{t+1}\mid Z_t,A_t)
\right).
\end{align}
Since the KL divergence is non-negative, the conditional
negative log-likelihood upper-bounds
$H(Z_{t+1}\mid Z_t,A_t)$. Therefore, minimising the predictive MSE minimises a variational upper bound on the conditional entropy.

Second, SIGReg encourages the aggregated latent distribution to match an isotropic Gaussian $q(Z_{t+1}) \approx \mathcal{N}(0,I)$. It is well known that, among all multivariate continuous distributions with a fixed covariance matrix, the Gaussian distribution maximises the Shannon differential entropy. %
Hence, SIGReg also maintains a high marginal entropy $H(Z_{t+1})$.
}

Combining these two effects, we can directly relate the objectives of the original LeWM as follows:
\begin{equation}
\underbrace{
\mathcal{L}_{\mathrm{pred}}
}_{\downarrow\, H(Z_{t+1}\mid Z_t,A_t)}
\;+\;
\lambda
\underbrace{
\mathcal{L}_{\mathrm{SIGReg}}
}_{\uparrow\, H(Z_{t+1})}.
\end{equation}
and since
\begin{equation}
I(Z_{t+1};Z_t,A_t)
=
H(Z_{t+1})
-
H(Z_{t+1}\mid Z_t,A_t),
\end{equation}
we arrive at the insight that the two objectives jointly optimise
\begin{equation}
\max\;
I(Z_{t+1};Z_t,A_t).
\end{equation}
Therefore, LeWM can be interpreted as a form of \emph{joint mutual information maximisation}.

{
$\Sigma$-LeWM's main modification lies in the backward and inverse directions: it optimises backward and inverse objectives directly through DCL, encouraging high mutual information between $Z_t$ and $(Z_{t+1},\tilde A_t)$, and between $\tilde A_t$ and $(Z_t,Z_{t+1})$, respectively.
}

In a similar fashion, the backward and inverse objectives directly maximise contrastive lower bounds on the corresponding mutual information terms:
\begin{subequations}
\begin{align}
    \mathcal{I}_{\mathrm{back}}
    &= I(Z_t; Z_{t+1}, \widetilde{A}_t),\\
    \mathcal{I}_{\mathrm{inv}}
    &= I(\widetilde{A}_t; Z_t, Z_{t+1}).
\end{align}
\end{subequations}
Specifically, we optimise both terms using DCL, which is monotonically related to the corresponding InfoNCE objective and therefore serves as a surrogate for maximising its mutual information lower bound.

Notably, we set out to optimise the sum of the three joint mutual information terms given above. 
It has been shown that such a sum directly relates to the Total Correlation (TC)~\citep{5392532} and Dual Total Correlation (DTC)~\citep{Han1978}, i.e.
\begin{equation}
    \begin{aligned}
    &\mathcal I_{\mathrm{forward}}
    + \mathcal I_{\mathrm{back.}}
    + \mathcal I_{\mathrm{inv.}}\\
    &=
    \mathrm{TC}(Z_t,\Tilde{A}_t,Z_{t+1})
    +
    \mathrm{DTC}(Z_t,\Tilde{A}_t,Z_{t+1}).
\end{aligned}
\label{eq:sum_TC_DTC}
\end{equation}

The decomposition in Eq.~(\ref{eq:sum_TC_DTC}) provides a useful information-theoretic interpretation of the proposed objective. TC quantifies the overall statistical dependence among $Z_t$, $\tilde{A}_t$, and $Z_{t+1}$, vanishing only when the three variables are mutually independent. DTC, by contrast, captures a complementary form of multivariate dependence that reflects how strongly each variable is constrained by the remaining variables. More importantly, their sum admits a particularly intuitive form:
\begin{equation}
\begin{aligned}
\Sigma(Z_t,\tilde{A}_t,Z_{t+1})
&=
\mathrm{TC}(Z_t,\tilde{A}_t,Z_{t+1})
+
\mathrm{DTC}(Z_t,\tilde{A}_t,Z_{t+1}),
\end{aligned}
\end{equation}
which is known as the S-Information~\citep{PhysRevE.100.032305}.

Thus, rather than encouraging only forward predictability,
our objective promotes predictive dependence in all three directions: the next state from the current state and action, the current state from the next state and action, and the action from the two consecutive states. We therefore refer to the proposed method as $\Sigma$-LeWM.

\section{Experiments}
\label{sec:exp}

\begin{figure}[t]
    \centering
    \includegraphics[width=0.8\columnwidth]{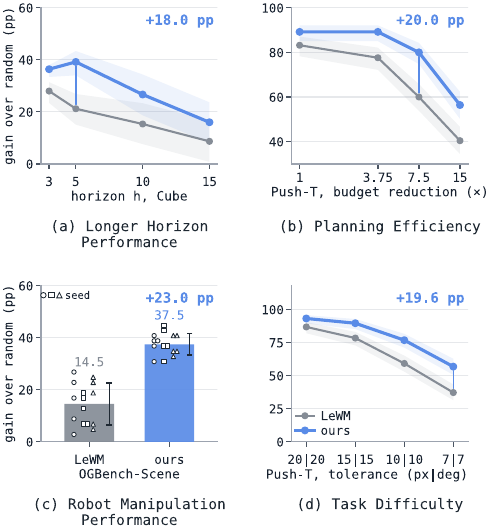}
    \caption{Overview of main experimental results: $\Sigma$-LeWM outperforms standard LeWM when considering long horizon tasks (a), and leads to improved planning efficiency (b). Both properties are important for complex and realistic robotic tasks. We can demonstrate superior performance over LeWM on the OGBench-Scene manipulation task (c; gain over the random-action floor; points: individual evaluations, marker shape: training seed, error bar: std over all 15 evaluations as in Table~\ref{tab:main}) and demonstrate better performance of $\Sigma$-LeWM when considering high-accuracy tasks in robotics (d).}
    \label{fig:placeholder}
    \label{fig:overview-exp}
\end{figure}

In our experiments, we set out to investigate whether (RQ1) the proposed $\Sigma$-LeWM shows performance gains over LeWM, (RQ2) what effect the auxiliary tasks have on the latent embedding and its structure, and, finally, (RQ3) if maximising the S-Information provides an advantage over maximising the individual terms. 

To answer the first question, we evaluate our method's predictive capacity under standard world model benchmarks in Section~\ref{subsec:standard}.
We also show the performance gain under longer planning horizons and lower compute effort, which is crucial in robotic scenarios, in Section~\ref{subsec:horizon}.
Furthermore, we introduce two more realistic robot manipulation tasks (Section~\ref{subsec:new_task}) to evaluate \ours{}'s end-to-end capability on more challenging tasks.

To answer (RQ2)
we perform dimensionality reduction on the learned latent space of $\Sigma$-LeWM and investigate whether the physical quantities of the task (e.g., object position) are preserved in the structure of the embedding (Section~\ref{subsec:latent}).
We also learn a GC-IDM~\citep{Nguyen2026} planner for both LeWM and \ours{}, evaluating whether the learned latent of \ours{} is easier for other planner methods to perform.

Finally, for (RQ3), we conduct a factorial ablation study on the two auxiliary tasks of backward and inverse dynamics, illustrating that only combining both, i.e., maximising the S-Information, leads to better task performance (Section~\ref{subsec:ablation}).

Figure \ref{fig:overview-exp} gives a short overview of the main experimental results, highlighting the performance gains and $\Sigma$-LeWM's increased ability to handle high-difficulty tasks common in robotics. 

\subsection{Experimental Setup}
\label{subsec:setup}
Across the tasks, we use the same set of auxiliary training parameters in the results unless mentioned, including the weight of the multi-step forward prediction $\lambda_{\mathrm{multi}}=1.0$ with maximum prediction offset $m=5$, and the weights with respect to the auxiliary objectives across all the experiments: $\lambda_{\mathrm{back}}=0.1$ for the backward, and $\lambda_{\mathrm{inv}}=0.05$ for the inverse; an ablation study on these parameters is shown in Section~\ref{subsec:ablation}.
We use DCL with temperature 0.1 and in-batch negatives (batch size 128).
Consistent with LeWM and to be comparable, we use the same encoder, predictor, optimiser, training protocols, and set of parameters related to the forward prediction training (learning rate, SIGReg weight $\lambda_{\mathrm{sig}}=0.09$, temperature of the SIGReg $\tau=0.1$) and the planning scheme using a CEM solver sampling 300 action sequences per iteration and selecting the top 30 candidates to update the sampling distribution.

\subsection{Results on the Standard Benchmarks}
\label{subsec:standard}
\begin{table}[t]
  \centering
  \revcap
  \small
  \caption{\textbf{Standard benchmark success (\%).} All methods use the LeWM CEM planner setting ($h{=}5$, CEM30). 
  Our measurements are over 3 training seeds $\times$ 5 evaluation seeds $\times$ 50 episodes.
  $^{\ddagger}$One of three training seeds collapses (20.0); the other two reach 99.6.
  $^{\dagger}$LeWM trained with our multi-step forward term only ($\lambda_{\mathrm{back}}=\lambda_{\mathrm{inv}}=0$), which is the ``base'' arm of Section~\ref{subsec:ablation}.
  $^{\S}$\ours{} with the backward term aligned directly to the raw latent $z_t$ (no projection head).
  }
  \label{tab:main}
\resizebox{0.9\columnwidth}{!}{%
  \begin{tabular}{@{}lccccc@{}}
    \toprule
    & Push-T & TwoRoom & Reacher & Cube & Scene \\
    \midrule
    Random policy (floor)
      & 2.0\,\std{2.0}
      & 23.6\,\std{3.6}
      & 13.6\,\std{3.8}
      & 43.6\,\std{8.9}
      & 43.2\,\std{6.7} \\

    LeWM \citep{Maes2026}
      & 90.0\,\std{5.7}
      & 90.9\,\std{4.1}
      & \textbf{86.0}\,\std{6.3}
      & 66.5\,\std{6.3}
      & 57.7\,\std{8.0} \\

    LeWM + multi-step$^{\dagger}$
      & 88.9\,\std{3.3}
      & 92.8\,\std{4.2}
      & 83.6\,\std{4.4}
      & 71.0\,\std{5.3}
      & 53.6\,\std{5.9} \\

    Sub-JEPA \citep{Zhao2026}
      & 84.0\,\std{8.7}
      & \textbf{93.6}\,\std{2.8}
      & 80.7\,\std{5.9}
      & 70.3\,\std{6.6}
      & 55.5\,\std{8.1} \\

    RC-aux \citep{Li2026}
      & 87.9\,\std{5.2}
      & 91.6\,\std{3.4}
      & 84.8\,\std{5.4}
      & 70.3\,\std{5.4}
      & 58.0\,\std{5.5} \\

    SMWM \citep{Ivashkov2026}
      & 88.5\,\std{5.4}
      & 73.1$^{\ddagger}$\,\std{38.9}
      & 84.5\,\std{5.8}
      & 81.9\,\std{4.9}
      & 60.0\,\std{3.1} \\

    AC-MTM \citep{Boylan2026}
      & 89.2\,\std{3.8}
      & 92.4\,\std{2.7}
      & 84.9\,\std{2.4}
      & 77.3\,\std{5.9}
      & 60.9\,\std{5.1} \\

    \revmark Backward to raw $z_t$$^{\S}$
      & \revmark 81.7\,\std{6.7}
      & \revmark 87.3\,\std{5.6}
      & \revmark 83.3\,\std{5.4}
      & \revmark 71.7\,\std{7.0}
      & \revmark 76.3\,\std{4.1} \\

    \rowcolor{oursbg}
    \ours{} (ours)
      & \textbf{94.5}\,\std{3.0}
      & 92.4\,\std{4.4}
      & \textbf{86.0}\,\std{4.8}
      & \textbf{82.6}\,\std{5.0}
      & \textbf{80.7}\,\std{4.2} \\

    \bottomrule
  \end{tabular}
  }
\end{table}
First, we evaluate our method and baselines on LeWM's four standard benchmarks and OGBench-Scene.

\textbf{Task setup.}
We use the standard Push-T, TwoRoom, Reacher, OGBench-Cube, and OGBench-Scene~\citep{Park2024} tasks and use default training and planning settings in LeWM.
We follow LeWM's default test-time planning settings: a CEM solver with 30 optimisation steps, a 5-step planning horizon, and environment timesteps set to 25 with a 5-step frame skip.
We compare against LeWM, Sub-JEPA~\citep{Zhao2026} which replaces SIGReg with a subspace Gaussian regulariser, RC-aux~\citep{Li2026}, which uses two auxiliary objectives for time/space alignment, and SMWM~\citep{Ivashkov2026} and AC-MTM~\citep{Boylan2026}, which add an inverse dynamics loss, in contrastive form for AC-MTM.
For LeWM and Sub-JEPA, training seed 3072 is the authors' released checkpoint and seeds 1 and 2 are retrained with the released recipe; RC-aux applies the authors' fine-tuning recipe to each LeWM seed. All baselines keep their original objectives and therefore train without our multi-step forward term. On Scene, which has no released checkpoints, we train every method ourselves.
To separate the contribution of the multi-step term from that of the backward and inverse objectives, Table~\ref{tab:main} also reports LeWM trained with the multi-step term alone on all five tasks (the ``base'' arm of Section~\ref{subsec:ablation}).
We report the mean and standard deviation over 3 training seeds $\times$ 5 evaluation seeds $\times$ 50 episodes.

\textbf{Results.}
Experiments (Table \ref{tab:main}) show that \ours{} is best on Scene by about 20 points and on Push-T by 4.5 points; on the other three tasks it is within noise of the best method: TwoRoom (92.4 vs.\ 93.6 for Sub-JEPA), Cube (82.6 vs.\ 81.9 for SMWM) and Reacher (86.0 vs.\ 86.0 for LeWM).
The multi-step forward term alone stays close to LeWM (Push-T 88.9, TwoRoom 92.8, Reacher 83.6, Cube 71.0, Scene 53.6), so the gains of \ours{} on Scene (80.7) and Cube come from the backward and inverse objectives.

It is worth noting that TwoRoom is a relatively low-complexity task, in which a point-mass has to navigate a gap. %

\subsection{Results on Extended Horizons}
\label{subsec:horizon}
Many real-world tasks, especially in robotics, benefit from long planning horizons. 
In these experiments we analyse further \ours{}'s behaviour under long-horizon planning.

\begin{figure}[t]
    \centering
    \includegraphics[width=0.8\columnwidth]{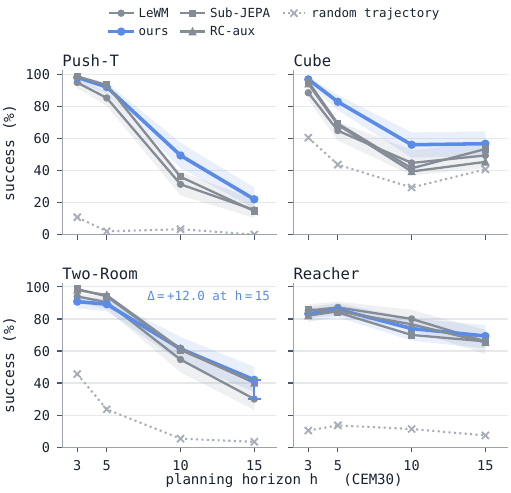}
    \caption{\textbf{Longer horizons}, at fixed inference effort CEM30 on all four tasks.
    $\Delta$: the performance gain over LeWM.
    }
    \label{fig:horizon}
\end{figure}

\textbf{Task setup.}
We extend the planning horizon from $h=5$ to $h=3,5,10,15$ to test robustness to longer lookahead planning, which is especially crucial for robot learning and planning.
We compare against LeWM, Sub-JEPA, and RC-aux with $h=3,5,10,15$ horizons using the four standard benchmarks above.

\textbf{Results.}
We show performance gains over all baselines on Push-T and Cube tasks across horizon choices, as shown in Fig. \ref{fig:horizon}.
Notably, we can observe a performance gain when extending the horizon to $h=15$ on TwoRoom and Reacher tasks, demonstrating the long-term planning capability of our method, which is essential for robot learning tasks.

\subsection{Results on Planning Efficiency}
\label{subsec:efficiency}
Planning with less compute is another crucial aspect for robot learning and planning.

\begin{figure}[t]
    \centering
    \includegraphics[width=0.8\columnwidth]{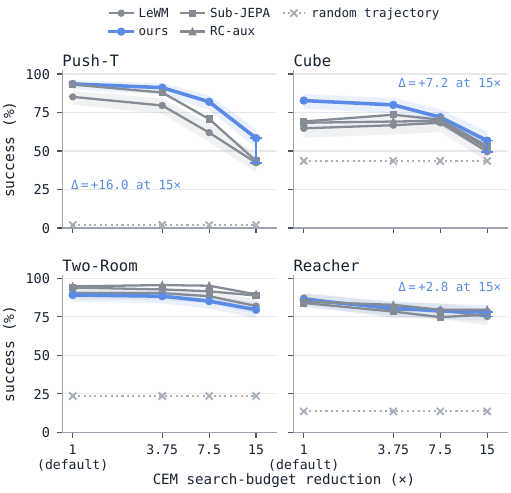}
    \caption{\textbf{Planning efficiency} with planning horizon $h{=}5$; the $x$-axis is the search reduction relative to LeWM's default planning budget CEM30. Moving right corresponds to lower test-time planning compute. 
  }
    \label{fig:efficiency}
\end{figure}

\textbf{Task setup.}
We therefore arrange a planning efficiency experiment with the CEM solver, changing its budget during test-time inference, where the number of iteration steps changes to $n_{\mathrm{CEM}}=8,4,2$ and therefore the CEM search budget is reduced by $\times3.75,\times7.5$ and $\times15$ (the $x$-axis of Figure~\ref{fig:efficiency}) compared to the standard $n_{\mathrm{CEM}}=30$.
We compare against the same baselines in Section~\ref{subsec:horizon}.

\begin{figure}[t]
    \centering
    \includegraphics[width=0.8\columnwidth]{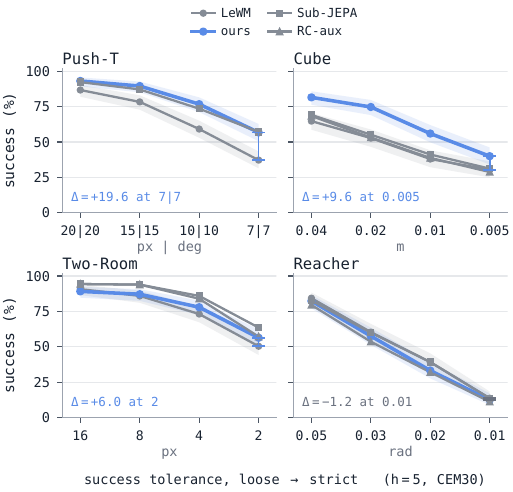}
    \caption{\textbf{Task difficulty.} $\Sigma$-LeWM outperforms LeWM under increasing task difficulty, where the success criterion is increasingly narrowed for goals (x-axis).
  }
    \label{fig:tolerance}
    \vspace{-3ex}
\end{figure}

\textbf{Results.}
We show in Fig. \ref{fig:efficiency} that our method outperforms the baselines with substantially less search on Push-T and Cube.
On Push-T, our method uses 8 CEM iterations to outperform LeWM at 30 CEM iterations (91.2\% vs.\ 85.2\%), and only 4 CEM iterations are already better than LeWM at 8 (Figure~\ref{fig:efficiency}) and remain competitive with LeWM at 30 while using a $\times7.5$ smaller CEM search budget.
Thus the gain is not only in the final success rate, but also in the test-time planning compute required to reach it.

\subsection{Robotic Tasks: Gear-Assembly and Cube-Stacking}
\label{subsec:new_task}
To evaluate if our method has an advantage on more realistic tasks, we introduce two manipulation tasks with strict success criteria.

\begin{figure}[t]
    \centering
    \revcap
    \revframe{\includegraphics[width=0.92\textwidth]{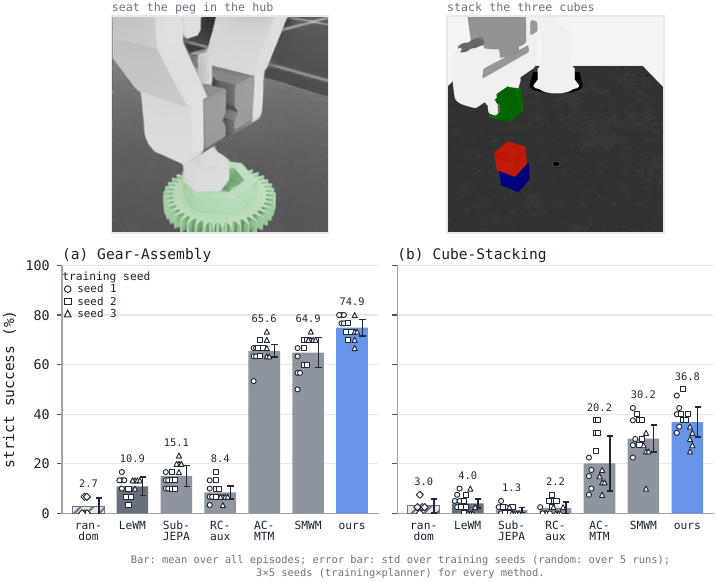}}
    \caption{
    Comparison of $\Sigma$-LeWM (ours) versus the baselines on robotic manipulation tasks: Gear-Assembly and Cube-Stacking. 
  }
    \label{fig:newtasks}
    \vspace{-3ex}
\end{figure}

\textbf{Task setup.}
We arrange the Cube-Stacking and Gear-Assembly tasks based on IsaacSim simulation environments~\citep{10107764} shown in Figure~\ref{fig:newtasks}.
The Cube-Stacking benchmark is derived from augmented teleoperation trajectories~\citep{pmlr-v229-mandlekar23a} (1k trajectories).
The Gear-Assembly dataset is collected by a learned policy from SRSA~\citep{Guo2025} with 10k trajectories, each with 75 environment steps.
We further filter the trajectories for strict success in terms of assembly and reduce the dataset to around 21k frames.
We use $\lambda_{\mathrm{sig}}=0$ for both Gear-Assembly and Cube-Stacking, and the others remain the same.
We compare against LeWM, Sub-JEPA, RC-aux, and baselines with additional dynamics terms, where SMWM~\citep{Ivashkov2026} utilises an inverse dynamics loss to avoid latent collapse, and AC-MTM~\citep{Boylan2026} shares the same inspiration but uses a contrastive version of it.
At test time, we use the first frame that satisfies the task-success criterion as the goal frame from the test set whose episodes are excluded from training, rather than relying on the fixed look-ahead goals used in the standard benchmarks.
We use horizon $h=5$ for Gear-Assembly and $h=10$ for Cube-Stacking.

\textbf{Results.}
We demonstrate (Fig.~\ref{fig:newtasks}) that our method improves strict success from 10.9\% for LeWM to 74.9\%, and also outperforms the inverse-dynamics baselines on Gear-Assembly.
Section~\ref{subsec:latent} examines one candidate explanation: whether the task-success state is linearly decodable from the frozen representation.

\subsection{Quality of the Latent Representation}
\label{subsec:latent}
We conduct two different analyses to illuminate the quality and structure of the learned embedding by our method: physical interpretation, and planner learnability.

\begin{figure}[t]
    \centering
    \includegraphics[width=\columnwidth]{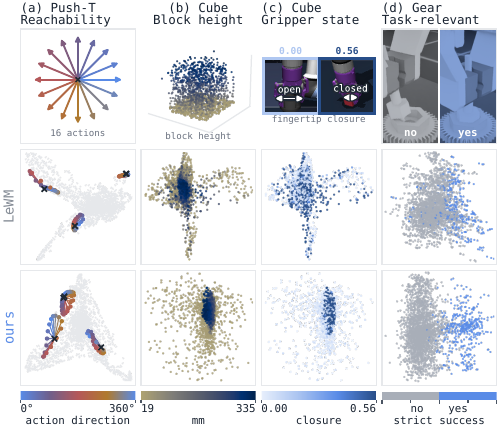}
    \caption{\textbf{Physical interpretation, LeWM vs \ours{}}. 
  PCA visualisations of each model's latent space, coloured by physical quantities of the tasks, showing explainable latent distributions.
  (a). Push-T 1-step reachability in the latent space from a fixed latent point $\mathbf{z}_t$ under sixteen normalised 2-D action directions, showing how far and in which directions $f(\mathbf{z}_t, \Tilde{\mathbf{a}_t})$ can move.
  (b). For OGBench-Cube block height, our representation shows a clear height-dependent structure, with high-block states concentrated in a compact region.
  (c). The gripper state in OGBench-Cube, with our representation showing closed-gripper states concentrated in a compact region, which LeWM separates.
  (d). For Gear, task-success states form a clearly separated latent region under ours, while LeWM shows a mixture between success and failure.
  }
    \label{fig:vis}
    \vspace{-3ex}

\end{figure}

\textbf{Physical interpretation.}
We show the principal component analysis (PCA) visualisation for the latent representation and highlight the specific physical quantities to illustrate if the learned latent reflects the distribution of a certain quantity.
In Figure~\ref{fig:vis}, the first four columns use PCA projections of the frozen latents, while Gear-Assembly in the last column uses a supervised linear view to distinguish whether the binary success signal is captured in the latent.
Each column represents the distribution of a certain physical quantity in different tasks, coloured by the value of the quantity.
Appendix~\ref{app:probes} quantifies them with held-out linear probes. 

\begin{figure}[t]
    \centering
    \includegraphics[width=0.85\columnwidth]{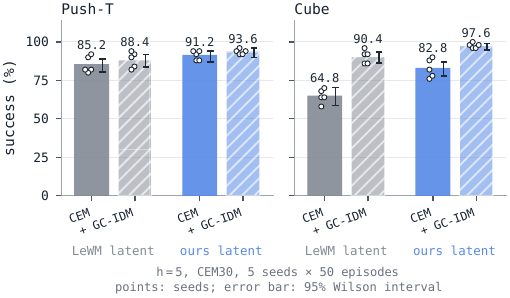}
    \caption{
    \textbf{Planner learnability, LeWM vs \ours{}.}
    Performance with default CEM planner and trained GC-IDM~\citep{Nguyen2026} planner on Push-T (left) and OGBench-Cube (right). The GC-IDM planners are trained on the learned latent representation with frozen encoders, and tested with 5 seeds $\times$ 50 episodes for each bin.
    }
    \label{fig:planner}

\end{figure}

\textbf{Planner learnability.}
\label{subsec:gcidm}
From the robot learning perspective, we also demonstrate the planner learnability on the latent representation to further evaluate more practically. 
Essentially, we illustrate that our method, combined with a state-of-the-art planner~\citep{Nguyen2026}, can further obtain performance gains compared to learning the planners on other latent representations.
We train the same GC-IDM~\citep{Nguyen2026} planner model separately on LeWM and our method and compare the OGBench-Cube performance.
We show in Figure~\ref{fig:planner} that with our latent representation, a learned planner performs better than the one on LeWM.

\subsection{Ablation Study: Why Backward/Inverse are Necessary}
\label{subsec:ablation}

\begin{figure}[t]
    \centering
    \includegraphics[width=0.85\columnwidth]{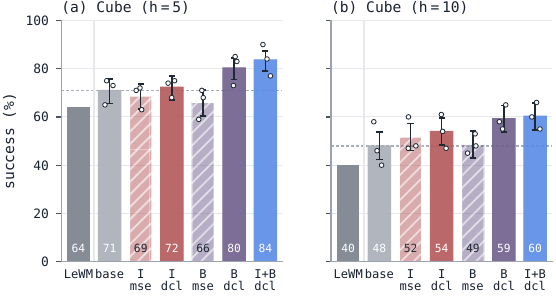}
    \caption{
    Comparison of $\Sigma$-LeWM versus standard LeWM and ablations of different combinations of loss functions. 
    On the x-axis, \emph{base} keeps only the multi-step forward objective ($\lambda_{\mathrm{back}}=\lambda_{\mathrm{inv}}=0$; dashed line); \emph{I} and \emph{B} add the inverse or the backward term alone, trained with a regression (\emph{mse}) or contrastive (\emph{dcl}) loss, and \emph{I+B dcl} is the full \ours{}.
    Evaluations with both short (h=5) and longer (h=10) horizon planning demonstrate the effectiveness and performance gain of $\Sigma$-LeWM when reflecting the S-Information.
    This ablation uses its own evaluation seeds (three banks of 100 episodes per checkpoint), so its values differ slightly from Table~\ref{tab:main}.
  }
    \label{fig:ablation}
\end{figure}

\textbf{Ablation on the objective types.}
We conduct factorial ablations on the effectiveness of the introduced contrastive tasks for the backward and inverse dynamics.
Specifically, we investigate the single-term effectiveness by keeping two terms of the proposed contrastive objective, or instead, the regression version.
We show in Figure~\ref{fig:ablation} that both backward and inverse objectives gain performance on OGBench-Cube, while combining them leads to the best performance.
In addition, using the regression objective (MSE) instead of the contrastive version in both cases results in a performance drop, indicating that the contrastive objectives capture better structure for the latent representation in terms of the final performance.

\begin{table}[t]
\centering\small
\setlength{\tabcolsep}{4pt}
\vspace{2ex}
\caption{Loss-weight ablation on Push-T: $h{=}5$ success (\%) when one term's weight is scaled while the other two stay frozen ($1\times$ are the weights used everywhere else in the paper).}
\begin{tabular}{lccc}
\toprule
loss weight multiplier & $0.5\times$ & $1\times$ & $2\times$ \\
\midrule
inverse $\LA$      & 82 & \textbf{90} & 87 \\
backward $\LB$     & 86 & \textbf{90} & 82 \\
SIGReg $\mathcal{L}_{\mathrm{SIG}}$ & 88 & \textbf{90} & 76 \\
\bottomrule
\end{tabular}
\label{tab:loss_weight}
\vspace{-3ex}
\end{table}

\textbf{Ablation on the objective weights.}
We scale one loss term for each training on Push-T with horizon $h{=}5$ while the other two remain as the default ($\lambda_{\mathrm{sig}}=0.09$, $\lambda_{\mathrm{back}}=0.1$, and $\lambda_{\mathrm{inv}}=0.05$).
We find that this set of weights is better in both scaling directions for every term (Table \ref{tab:loss_weight}). This sweep uses evaluation seeds different from those of Table~\ref{tab:main}.

\section{Conclusion}
We propose a new set of auxiliary tasks to enable the learning of robust and well-structured latent embeddings in world models such as LeWorldModel. 
Our proposed loss-functions are based on insight from information theory, and complement the standard objectives of LeWM in a theory-grounded way. 
Adding loss functions for backward and inverse dynamics in the latent space is equivalent to maximising the Dual and Total Correlations, in short the S-Information. 
This approach leads to an increase in determinism, or predictability, between the latent variables while rewarding their alignment at the same time. 
Experiments demonstrated that the proposed method $\Sigma$-LeWM shows strong performance improvements over LeWM, especially in robot manipulation benchmarks. 
Furthermore, we note an increased stability for long-horizon planning, an important feature for robotic tasks in the real world. 
Future work will investigate the use of the current auxiliary tasks, the backward and inverse latent dynamics, at planning time and the deployment of $\Sigma$-LeWM on real-world robot tasks.

\bibliographystyle{IEEEtran}
\bibliography{icra2027}

\clearpage
\appendix
\revmark\revcap
\ifreviewmark\renewcommand{\std}[1]{{\scriptsize$\pm$#1}}\fi
\makeatletter\ifreviewmark\g@addto@macro\@floatboxreset{\revmark}\fi\makeatother
\section{Quantitative Probes of the Frozen Representation}
\label{app:probes}
To complement the qualitative PCA views of Figure~\ref{fig:vis}, we measure how much simulator state a linear probe recovers from the frozen latent representation on OGBench-Cube.

\textbf{Task setup.}
We freeze all the encoders, and all frames come from fresh episodes unseen by every encoder.
We fit ridge regression, or logistic regression for binary targets, on standardised features, using episode-grouped 60/20/20 splits that are shared by all encoders and repeated with five seeds.

\begin{table}[t]
\centering\small
\captionsetup{width=\textwidth}
\caption{\textbf{Linear probes on OGBench-Cube} (1800 frames from 300 unseen episodes). Test score, mean\,$\pm$\,sd over 5 episode-grouped splits.}
\label{tab:probe_cube}
\resizebox{\textwidth}{!}{%
\begin{tabular}{@{}lccccccc@{}}
\toprule
 & Block height $R^2$ & Gripper closed AUROC & Gripper closed Acc. & Gripper opening $R^2$ & Effector $xyz$ $R^2$ & Block $xy$ $R^2$ & Effector$-$block $R^2$ \\
\midrule
LeWM & 0.958\,\std{0.006} & \textbf{0.988}\,\std{0.005} & 0.950\,\std{0.012} & 0.837\,\std{0.035} & 0.837\,\std{0.023} & 0.929\,\std{0.030} & 0.775\,\std{0.055} \\
LeWM + multi-step & 0.951\,\std{0.012} & 0.986\,\std{0.004} & 0.951\,\std{0.011} & 0.838\,\std{0.026} & 0.980\,\std{0.000} & 0.647\,\std{0.037} & 0.728\,\std{0.025} \\
Sub-JEPA & 0.755\,\std{0.022} & 0.980\,\std{0.005} & 0.944\,\std{0.005} & 0.761\,\std{0.026} & 0.929\,\std{0.005} & 0.898\,\std{0.013} & 0.887\,\std{0.006} \\
RC-aux & 0.958\,\std{0.005} & \textbf{0.988}\,\std{0.006} & 0.954\,\std{0.009} & 0.840\,\std{0.035} & 0.843\,\std{0.026} & \textbf{0.933}\,\std{0.026} & 0.783\,\std{0.062} \\
\rowcolor{oursbg}
\ours{} & \textbf{0.978}\,\std{0.005} & \textbf{0.988}\,\std{0.007} & \textbf{0.963}\,\std{0.014} & \textbf{0.888}\,\std{0.027} & \textbf{0.997}\,\std{0.000} & 0.846\,\std{0.016} & \textbf{0.900}\,\std{0.014} \\
\bottomrule
\end{tabular}}
\end{table}

\textbf{Results.}
We find that the two terms play different roles (Table~\ref{tab:probe_cube}).
We observe that the multi-step forward term improves the robot's own state, increasing the end-effector accuracy from $0.84$ to $0.98$, but that it does so at the cost of object information: the block's planar position falls from $0.93$ to $0.65$.
However, adding the backward and inverse terms recovers most of this loss while preserving the improvement in end-effector representation. 
It also improves the representation of the relative position between the end-effector and the block compared with either alternative.
We also find that the gripper state is already saturated under LeWM.

\section{Measured Planning Latency}
\label{app:latency}
\newcommand{\latfixed}{7.1--9.3}
\newcommand{\latperiter}{22.1--23.2}
\newcommand{\latfixedmean}{8}
\newcommand{\latperitermean}{23}
\newcommand{\latpredshare}{98}
\newcommand{\latgpu}{GeForce RTX 3090}
\newcommand{\latsamples}{300}
\newcommand{\latplanshare}{41--74}
\newcommand{\latplansharelow}{11--32}
\newcommand{\latwallspeedup}{1.5--2.7}
\newcommand{\latgroup}[1]{\multicolumn{5}{@{}l}{\textit{#1}}}
\begin{table}[t]
\centering\small
\captionsetup{width=0.72\textwidth}
\caption{\textbf{Measured planning latency of the CEM ladder} for a single-environment MPC solve, $h{=}5$, \latsamples{} candidates, on two GPUs. Mean\,$\pm$\,sd over the four standard tasks and over LeWM, Sub-JEPA, RC-aux and \ours{}, whose planners are identical at test time. Replanning rate and speed-up include observation preprocessing (${\approx}1$\,ms); latency does not.}
\label{tab:latency}
\resizebox{0.72\textwidth}{!}{%
\begin{tabular}{@{}l@{\hspace{1.2em}}cccc@{}}
\toprule
 & CEM30 & CEM8 & CEM4 & CEM2 \\
\cmidrule(l){2-5}
Nominal budget reduction & $\times$1 & $\times$3.75 & $\times$7.5 & $\times$15 \\
\midrule
\latgroup{NVIDIA GeForce RTX 3090} \\
\quad Solve latency (ms) & 692.2\,\std{7.9} & 191.0\,\std{1.9} & 99.5\,\std{0.9} & 53.8\,\std{0.7} \\
\quad Replanning rate (Hz) & 1.44\,\std{0.02} & 5.21\,\std{0.05} & 9.97\,\std{0.09} & 18.31\,\std{0.22} \\
\quad Measured speed-up & $\times$1.00 & $\times$3.61\,\std{0.02} & $\times$6.91\,\std{0.05} & $\times$12.69\,\std{0.10} \\
\addlinespace[3pt]
\latgroup{NVIDIA H100} \\
\quad Solve latency (ms) & 450.9\,\std{3.6} & 127.0\,\std{1.2} & 68.2\,\std{0.7} & 38.6\,\std{0.6} \\
\quad Replanning rate (Hz) & 2.21\,\std{0.02} & 7.82\,\std{0.07} & 14.48\,\std{0.14} & 25.30\,\std{0.32} \\
\quad Measured speed-up & $\times$1.00 & $\times$3.53\,\std{0.01} & $\times$6.54\,\std{0.03} & $\times$11.43\,\std{0.08} \\
\bottomrule
\end{tabular}}
\end{table}

Section~\ref{subsec:efficiency} reduces the CEM search budget, which counts optimisation iterations rather than time; we therefore measure the latency of the planner itself.

\textbf{Task setup.}
We evaluate the inference time of the standard single-environment MPC solver on a single desktop \latgpu{} and, as a data centre reference, a single NVIDIA H100, with 5 warm-up solves and 50 repetitions per setting.

\textbf{Results.}
We find that solve time grows linearly with the number of CEM iterations, almost all of it spent by the predictor rolling out candidate plans (Table~\ref{tab:latency}).
We observe that cutting CEM30 to CEM4 makes a solve about seven times faster on both GPUs.
We also find that the H100 is only about $1.5{\times}$ faster than the \latgpu{}, since a ViT-tiny predictor scoring \latsamples{} candidates is latency-bound rather than compute-bound.
All four methods time within two percent of each other, because the auxiliary heads are discarded after training.

\end{document}